# MVDepthNet: Real-time Multiview Depth Estimation Neural Network

Kaixuan Wang Shaojie Shen
Hong Kong University of Science and Technology
Hong Kong
kwangap@ust.hk eeshaojie@ust.hk

## Abstract

*Although deep neural networks have been widely applied to computer vision problems, extending them into multiview depth estimation is non-trivial. In this paper, we present MVDepthNet, a convolutional network to solve the depth estimation problem given several image-pose pairs from a localized monocular camera in neighbor viewpoints. Multiview observations are encoded in a cost volume and then combined with the reference image to estimate the depth map using an encoder-decoder network. By encoding the information from multiview observations into the cost volume, our method achieves real-time performance and the flexibility of traditional methods that can be applied regardless of the camera intrinsic parameters and the number of images. Geometric data augmentation is used to train MVDepthNet. We further apply MVDepthNet in a monocular dense mapping system that continuously estimates depth maps using a single localized moving camera. Experiments show that our method can generate depth maps efficiently and precisely.*

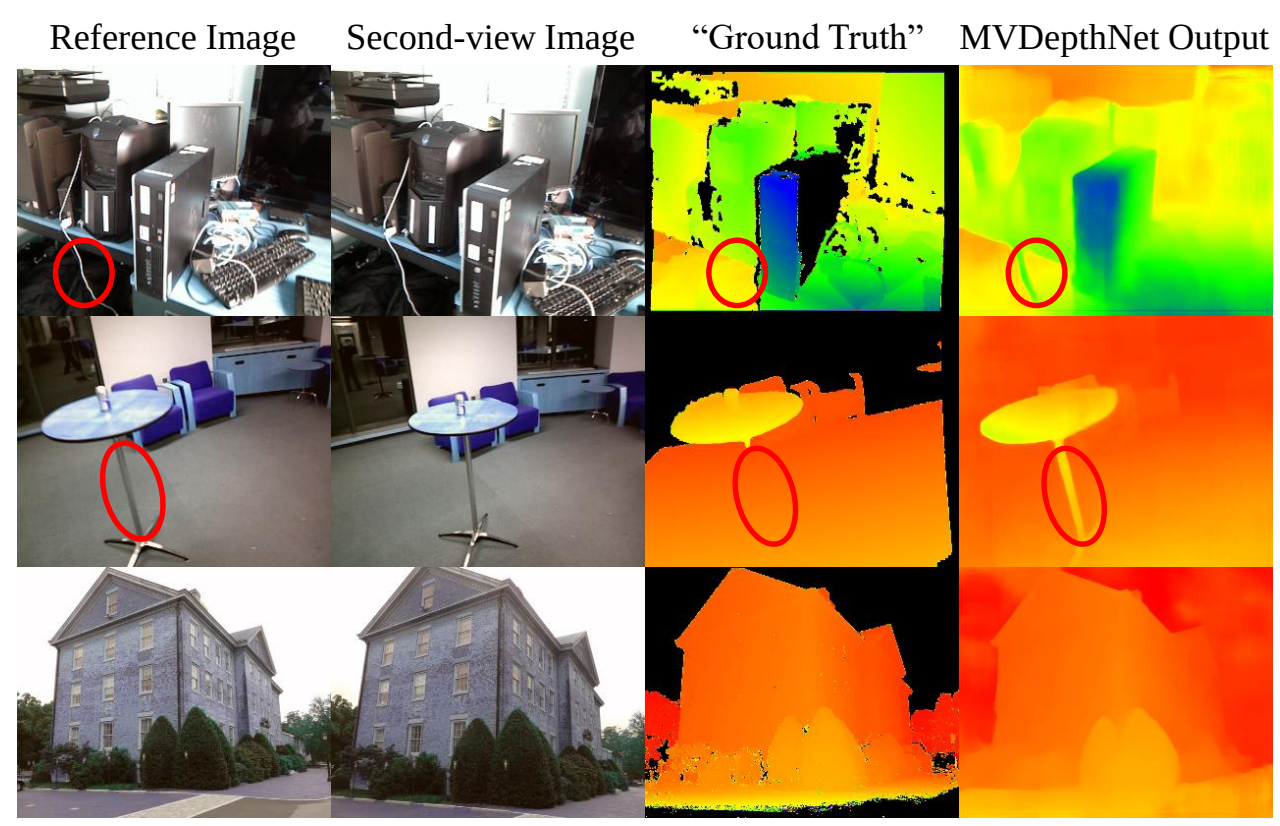

Figure 1. Examples to show the performance of MVDepthNet. MVDepthNet takes two or more image-pose pairs from a localized monocular camera as the input and estimates the depth map of the reference image. “Ground Truth” is provided by the dataset using RGB-D cameras or offline multiview stereo methods. As shown, MVDepthNet can not only estimate smooth and accurate depth maps but also handle situations that are difficult for RGB-D cameras, such as tiny structures and reflective surfaces, as marked in red circles. Also, it can generate large distance estimations, as shown in the third row.

## 1. Introduction

Depth estimation from images is an important task in computer vision and robotics. Stereo cameras are widely used in robotic systems to get the depth estimation. The depth of a pixel can be triangulated easily if its correspondence can be found along the same row in the other image. Although stereo systems achieve impressive results, monocular systems are more suitable for robots, especially micro robots, to estimate the dense depth maps for a number of reasons. First, to perceive the environment in different views, stereo systems need a baseline space from several centimeters (*e.g.*, $10\ cm$ in VI-Sensor [1]) to several decimeters (*e.g.*, $54\ cm$ in kitti setup [2]). On the other hand, monocular systems consist of a single camera and are much smaller. Second, the estimated depth range of monocular systems is not limited because the observation baseline is not fixed. Large baseline observations can be utilized to triangulate distant objects, and short baseline observations can be used to increase the robustness to repeated patterns. Third, recently developed visual odometry and visual-inertial systems, such as ORB-SLAM [3] and OKVIS [4], can provide monocular depth estimation with high-precision camera poses in real-time. Visual-inertial system [5] reports high accuracy poses estimation with only $0.88\%$ drift in large-scale environments. On the contrary, stereo systems need to be calibrated offline and maintained carefully during the runtime.

Many methods have been proposed to solve the depth estimation problem using a single camera. These methods can be divided into two main categories: multiview stereopsis-based estimation, and one image depth prediction. Traditional multiview-based approaches (*e.g.*, DTAM [6]) feature a standard pipeline including matching cost computation, cost aggregation, depth estimation and depth refinement. The handcrafted pipeline makes these methods hard to in-corporate more depth cues such as shading and structure priors. DeMoN [7] and DeepMVS [8] estimate depth maps from two-view or multi-view images using convolutional neural networks (CNNs). Because DeMoN [7] and DeepMVS [8] consist of several stacked networks, they are not computationally efficient for robotic applications. Depth prediction methods (*e.g*., [9], and [10]) usually use CNNs to learn the relationship between depth maps and image contents. The depth prediction relies on the prior knowledge from the training dataset and does not require any multiview observations. However, the network generalization ability and the scale ambiguity from a single image limit the application of depth prediction methods.

In this paper, we focus on solving multiview depth estimation given images and camera poses from a moving localized monocular camera by an end-to-end trained CNN. Such a network can exploit the information in multiview observations and can estimate depth maps in real-time. Although CNNs have been successfully used in many areas in computer vision including stereo matching (*e.g*., CRL [11] and GC-NET [12]) and optical flow (*e.g*., FlowNet [13] and FlowNet 2.0 [14]), extending CNNs into multiview depth estimation is challenging. First, depth is estimated by searching the corresponding pixels along epipolar lines in other images taken from different viewpoints. Unlike stereo matching, where the search line is fixed and can be encoded into the network architecture, the search line is arbitrary in monocular systems determined by the camera intrinsic parameters and camera poses. Searching pixels without epipolar constraints may be a solution but it comes at the cost of large search space and inaccurate results. Second, in monocular cases, the triangulation of pixel pairs depends on pixels' locations, camera intrinsic parameters, and relative poses. Parameterization of the information into the network is non-trivial. Third, since the location of every pixel is constrained by multiview geometry, traditional data augmentation like image scale and flip cannot be directly used to train the network.

To overcome these challenges, we propose a network to extract depth maps from a constructed cost volume which contains the multiview observations from input images. Since the multiview information is converted into the cost volume, the network does not need to match pixels across images and can be applied regardless of the camera types and the number of images. Estimating depth maps directly from the cost volume enables the network to achieve high-efficiency that can be used in real-time systems. To train the network using limited datasets, we further propose geometric data augmentation. Instead of transforming the input images, the augmentation is applied to the cost volume and the corresponding ground truth depth image so that per-pixel consistency is maintained. We apply our network into a real-time monocular dense depth estimation system and compare it with a variety of state-of-the-art methods. Figure 1 shows the performance of our network in indoor and outdoor environments. As shown in the figure, the network correctly estimates the depth of objects that are difficult for RGB-D cameras to perceive or are far away from the camera.

## 2. Related Work

A variety of works have been proposed to estimate the depth using a single camera based on multiview stereopsis or single view depth prediction.

DTAM [6] and REMODE [15] are two methods that extract depth maps using multiview images. DTAM [6] extracts the depth map from a multiview cost volume by minimizing an energy function considering both the matching cost and the depth map smoothness. REMODE [15] estimates the depth of pixels by searching their correspondences in other images and updates a robust probabilistic model. Finally, depth maps are smoothed by a total variation optimization.

DeMoN [7] is a learning-based method that takes two images as the input and estimates the depth and motion. To find pixel correspondences between images, the network firstly estimates the optical flow and then converts it into depth and ego-motion estimation. Multiple stacked encoder-decoders are used in the network to refine the results. Although it outperforms the baseline method [16] on a variety of datasets, DeMoN [7] has the following limitations. First, the camera intrinsic parameters are fixed in the network. Images with a large field of view (FOV) need to be cropped before the process, leading to a loss of view angle. Second, the network cannot be extended easily to more than two images.

DeepMVS [8] poses the depth estimation as a multi-class classification problem and can be applied to multiple input images. Image patches are matched between input images using patch matching networks and the results are aggregated intra-volume. A max-pooling layer is used to handle cost volumes from multiple image pairs and then the cost is aggregated inter-volume to give out the final prediction. To further improve the quality, the Dense-CRF [17] is used to smooth the depth map. Since the matching cost of every image patch needs to be processed by a deep network and the Dense-CRF [17] is used to refine the depth estimation, DeepMVS [8] cannot estimate depth maps in real-time.

Recently, many learning-based methods (*e.g*., [9], [10], and [18]) have shown good performance to predict the depth of an image. Tateno *et al*. [19] further show that the predicted depth estimation can be used to solve SLAM problems by proposing CNN-SLAM. Although they demonstrate good performance in specific environments, the application domain of these methods is limited by the training dataset. The scale ambiguity is another reason that limits the usability.

Here, we show that depth maps can be directly estimated given multiple images and the corresponding camera poses using a neural network without any post-process. The proposed method is trained end-to-end and generates high-quality depth maps in real-time. Thanks to the simple but effective structure of the network, our method can process multiple images from larger FOV cameras.

## 3. System Overview

The proposed neural network takes multiple images with known camera intrinsic parameters and poses as the input and generates an inverse depth estimation of the reference image. We call our multiview depth estimation network *MVDepthNet*. Input images are firstly converted into a cost volume, where each element records the observation of a pixel in different views at a certain distance. An encoder-decoder network is then used to extract the inverse depth maps using the constructed cost volume and the reference image. Our method can be easily extended to different camera configurations and applied to real-time monocular depth estimation using a single moving camera.

The network architecture is introduced in Section 4. In Section 5, the trained network is then applied into a monocular dense mapping system that continuously generates depth maps using a single localized camera. We compare our method with a variety of state-of-the-art methods in Section 7 to demonstrate the performance.

## 4. MVDepthNet

The structure of MVDepthNet is largely inspired by classic multiview stereo in which the depth is solved given the pixel matching cost volume and the reference image. However, unlike classic methods which consist of several sperate steps, MVDepthNet is designed to directly learn the relationship between depth maps and the given information from a large dataset.

### 4.1. Preliminaries

The input of the system is several image-pose pairs $\{I_i, \mathbf{T}_{w,i}\}$, where $I_i$ is the input RGB image and $\mathbf{T}_{w,i}$ is the pose of the camera frame with respect to the world frame $w$ when taking $I_i$. $\mathbf{T}_{w,i} \in \mathbb{SE}(3)$ consists of a rotation matrix $\mathbf{R}_{w,i} \in \mathbb{SO}(3)$ and a translation vector $\mathbf{t}_{w,i}$:

$$\mathbf{T}_{w,i} = \begin{bmatrix} \mathbf{R}_{w,i} & \mathbf{t}_{w,i} \\ \mathbf{0}^T & 1 \end{bmatrix}. \quad (1)$$

The image to estimate the depth map is called the reference image, and other images are called measurement images. A 3D point in the camera frame $\mathbf{x}_c = [x, y, z]^\mathrm{T}$ can be projected on the image as a pixel $\mathbf{u} := [u, v]^\mathrm{T}$ using the projection function: $\mathbf{u} = \pi(\mathbf{x}_c)$. Similarly, a pixel can be back-projected into the space as a 3D point $\mathbf{x}_c = \pi^{-1}(\mathbf{u}, d)$ where $d$ is the depth of the pixel $\mathbf{u}$. The projection function $\pi(\cdot)$ and the back-project function $\pi^{-1}(\cdot)$ are determined by camera models and intrinsic parameters which are known in the system. A 3D point in the camera frame $\mathbf{x}_c$ can be transformed into the world frame as $\mathbf{x}_w = \mathbf{T}_{w,i}\mathbf{x}_c$. Also, points in the world frame can be converted into the camera frame using $\mathbf{x}_c = \mathbf{T}_{w,i}^{-1}\mathbf{x}_w$.

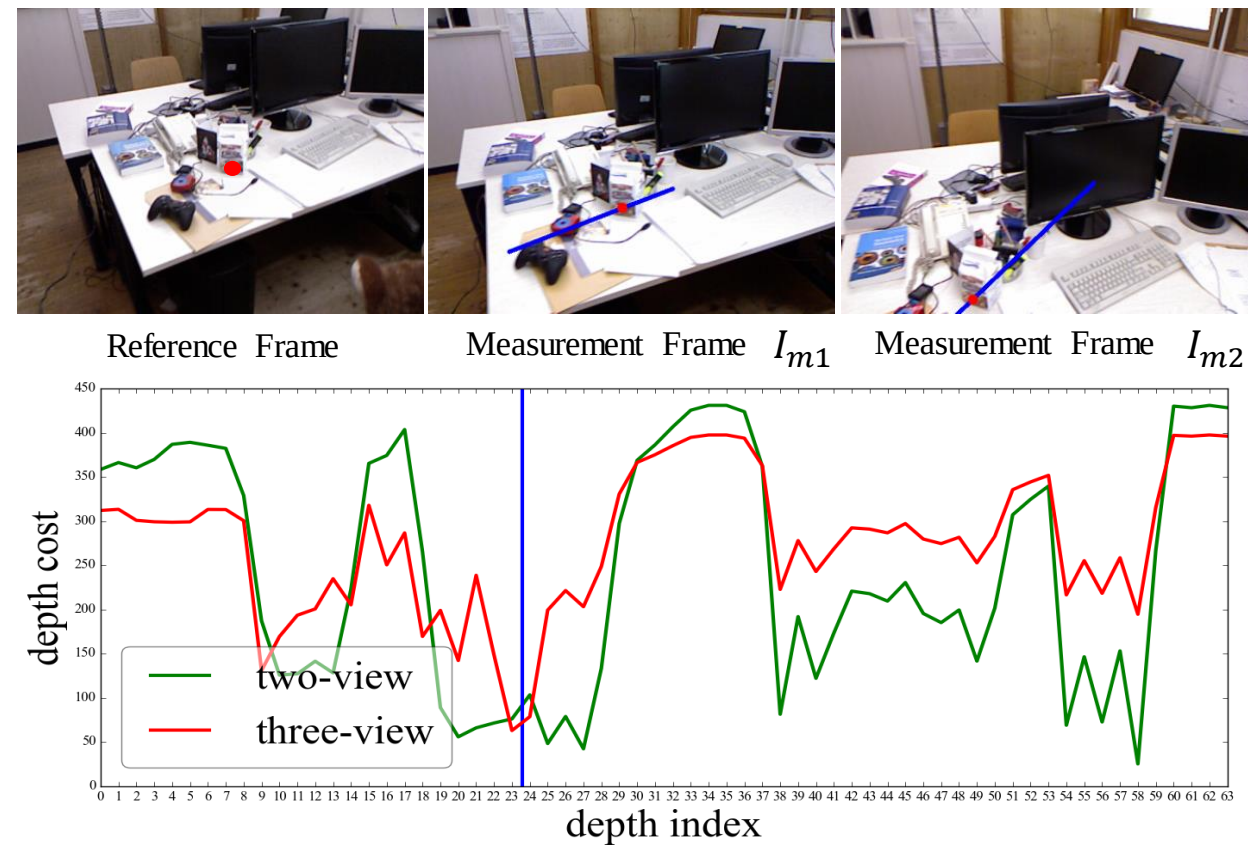

Figure 2. A visualization of fusing measurement frames into the cost volume. Top: the reference frame. Middle: two measurement frames used to calculate the cost volume. A pixel in the reference frame is highlighted in red to show the extracted multiview information. In the measurement frames, matched pixels during cost calculation are colored in blue and the true corresponding points are marked in red. Bottom: the cost of the pixel at different sampled depth values. The cost using only $I_{m1}$ is shown in green, and the cost using both $I_{m1}$ and $I_{m2}$ is shown in red. The blue line in the bottom of the figure is the ground truth depth. Both robustness and accuracy of the cost volume benefit from multiview observations.

### 4.2. Network Architecture

In optical flow estimation or stereo matching, the matching cost between image patches is important to find pixel correspondence. Traditional methods usually rely on handcrafted similarity metrics based on mutual information, cross-correlation or other functions. To deal with textureless region and occlusion, the cost volume is smoothed by specific filtering approaches, such as guided filter, adaptive guided filtering, etc. On the other hand, learning based methods (*e.g.*, FlowNet 2.0 [14], and CRL [11]) usually use 'correlation layers' to explicitly correlates the extracted feature maps of two images. Instead of using networks to extract feature descriptor for each pixel which is computationally expensive and memory intensive, we use the network to extract depth maps from constructed cost volumes by learning cost aggregation, depth estimation, and refinement directly from data without any handcrafted operation.

In our network, a 3D matching cost volume is built to represent the observation of pixels in other measurement images at different distances. The matching cost of pixel

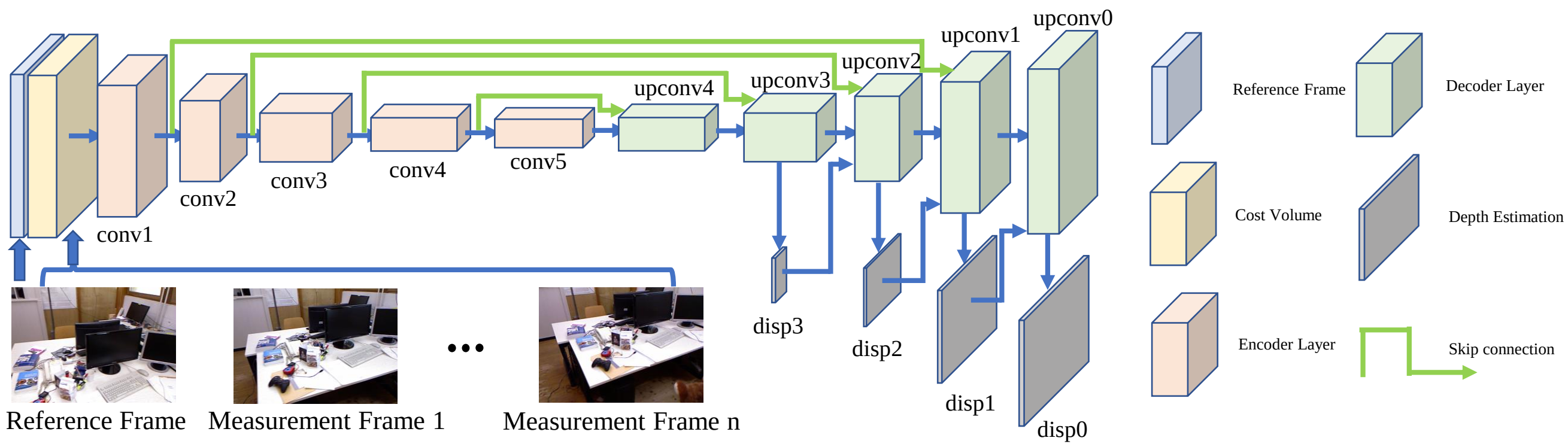

Figure 3. Overview of the architecture of MVDepthNet. Multiple measurement frames are encoded in the cost volume. MVDepthNet takes the reference frame and the cost volume as the input and estimates inverse depth maps at four resolutions. Skip connections between the encoder and decoder are used to keep the local information of pixels, which is important for the per-pixel depth estimation. Detailed definition is listed in Table 1.

| Name | Kernel | Stride | Ch I/O | Res I/O | Input |
|---|---|---|---|---|---|
| conv1 | 7x7 | 1 | 67/128 | 0/0 | reference image + cost volume |
| conv1_1 | 7x7 | 2 | 128/128 | 0/1 | conv1 |
| conv2 | 5x5 | 1 | 128/256 | 1/1 | conv1_1 |
| conv2_1 | 5x5 | 2 | 256/256 | 1/2 | conv2 |
| conv3 | 3x3 | 1 | 256/512 | 2/2 | conv2_1 |
| conv3_1 | 3x3 | 2 | 512/512 | 2/3 | conv3 |
| conv4 | 3x3 | 1 | 512/512 | 3/3 | conv3_1 |
| conv4_1 | 3x3 | 2 | 512/512 | 3/4 | conv4 |
| conv5 | 3x3 | 1 | 512/512 | 4/4 | conv4_1 |
| conv5_1 | 3x3 | 2 | 512/512 | 4/5 | conv5 |
| upconv4 | 3x3 | 1 | 512/512 | 4/4 | conv5_up |
| iconv4 | 3x3 | 1 | 1024/512 | 4/4 | upconv4 + conv4_1 |
| upconv3 | 3x3 | 1 | 512/512 | 3/3 | iconv4_up |
| iconv3 | 3x3 | 1 | 1024/512 | 3/3 | upconv3+conv3_1 |
| disp3 | 3x3 | 1 | 512/1 | 3/3 | iconv3 |
| upconv2 | 3x3 | 1 | 512/256 | 2/2 | iconv3_up |
| iconv2 | 3x3 | 1 | 513/256 | 2/2 | upconv2+conv2_1+disp3_up |
| disp2 | 3x3 | 1 | 256/1 | 2/2 | iconv2 |
| upconv1 | 3x3 | 1 | 256/128 | 1/1 | iconv2_up |
| iconv1 | 3x3 | 1 | 257/128 | 1/1 | upconv1+conv1_1+disp2_up |
| disp1 | 3x3 | 1 | 128/1 | 1/1 | iconv1 |
| upconv0 | 3x3 | 1 | 128/64 | 0/0 | iconv1_up |
| iconv0 | 3x3 | 1 | 65/64 | 0/0 | upconv0+disp1_up |
| disp0 | 3x3 | 1 | 64/1 | 0/0 | iconv0 |

Table 1. Details of MVDepthNet where 'Ch I/O' is the channel number of input/output features, 'Res I/O' is the downsampling factor of input/output layers and 'Input' is the input of the layers. All layers except 'disp*' are followed by batch normalization [20] and ReLU. '*_up' means the upsampled feature maps. These feature maps are upsampled using bilinear interpolation. 'disp*' is a convolution layer followed by a sigmoid function and scaled by two. '+' is the concatenation operation.

$\mathbf{u}_r$ of reference frame $I_r$ at sampled distance $d$ is defined as

$$C(\mathbf{u}_r, d) = \frac{1}{|\mathcal{M}|} \sum_{I_m \in \mathcal{M}} AD(I_m, I_r, \mathbf{u}_r, d), \qquad (2)$$

where $\mathcal{M}$ is the measurement images set. For every measurement image $I_m$ and distance $d$, the pixel $\mathbf{u}_r$ in reference image $I_r$ is projected to $I_m$ as $\mathbf{u}_m = \pi(\mathbf{T}_{w,m}^{-1} \mathbf{T}_{w,r} \pi^{-1}(\mathbf{u}, d))$. $AD(I_m, I_r, \mathbf{u}_r, d)$ calculates the absolute intensity difference between $\mathbf{u}_r$ on image $I_r$ and the corresponding pixel $\mathbf{u}_m$ on measurement frame $I_m$.

There are two advantages by using absolute difference as the matching cost. First, compared with complex similarity metrics, the per-pixel matching cost is much more efficient to calculate and extend to multiple measurement images. Second, the cost is calculated per-pixel without a supporting window so that detail information can be preserved which is important for the reconstruction of fine objects. $N_d$ depth values are uniformly sampled on inverse depth space from $d_{min}$ to $d_{max}$. The $i$-th sampled depth $d$ is given as

$$\frac{1}{d} = \left(\frac{1}{d_{min}} - \frac{1}{d_{max}}\right) \frac{i}{N_d - 1} + \frac{1}{d_{max}}. \qquad (3)$$

The information extracted from multiview images is shown in Figure 2. Although the network does not have direct accesses to multiview images, the cost volume effectively encodes the multiview constrains. As more measurement images are encoded into the cost volume, the matching cost of a pixel is more distinct and robust to occlusion.

The reference image stacked with the constructed cost volume is sent to MVDepthNet, which consists of an encoder-decoder architecture. The network is shown in Figure 3 and shares similar spirits with FlowNetSimple [13] but has several important adjustments. The encoder is used to extract global information of the image and aggregate high-level pixel costs. The extracted global and high-level information is then upsampled into finer resolutions by the decoder architecture. To keep the low-level per-pixel information, skip connections are used to combine low-level cost and high-level information. Thanks to the full resolution cost volume, MVDepthNet generalizes depth maps in four resolutions with the finest resolution the same as input images. Since the estimated inverse depth is bounded between 0 and $1/d_{min}$, a scaled sigmoid function is used to constrain the output range. The detailed definition of each layer of the MVDepthNet is listed in Table 1.

### 4.3. Geometric Data Augmentation

Data augmentation is used to train MVDepthNet. Apart from widely used image augmentation, such as adding noise, changing brightness, contrast, and color, two additional geometric augmentation methods are proposed for our task.

MVDepthNet is proposed to estimate the depth maps of images taken indoors, outdoors where the distance varies from several meters to tens of meters. However, most of the training data are recorded by RGB-D cameras that have a perception range of less than $8$ meters and only work indoors. Short-range supervision information will bias the network in that it prefers short distance estimation. To extend the depth range, we scale the depth value together with the translation of camera poses by a factor of between $0.5$ to $1.5$.

Image flip and scale are widely used in classification network training. However, images, camera intrinsic parameters, and the relative camera poses are constrained by multiview geometry. Instead of changing the intrinsic parameters and the relative poses according to geometry constraints before calculating the cost volume, we apply the transformation to constructed cost volumes, reference images and the corresponding ground truth depth maps. The geometric transformation includes random flip and scale of between $1.0$ and $1.2$.

# 5. Monocular Depth Estimation System

MVDepthNet is a network that takes several images as the input and generates one inverse depth map. We apply the trained MVDepthNet into a monocular depth estimation system that can continuously generate depth maps given image-pose sequences from a localized moving camera.

Measurement frames are selected to calculate the cost volume for each reference frame. A simple frame selection method is used to highlight the performance of MVDepthNet. The first image in the sequence is selected as the measurement frame. The $k$-th measurement frame is selected when its corresponding camera pose has enough view angle difference or baseline length compared with the former measurement frame. The relative rotation between frame $i$ and frame $j$ is $\mathbf{R}_{j,i} = \mathbf{R}_{w,j}^{-1}\mathbf{R}_{w,i}$ and the view angle difference is defined as

$$\theta_{j,i} = arccos((\mathbf{R}_{j,i}[0,0,1]^T) \cdot [0,0,1]). \tag{4}$$

$\theta_{j,i}$ measures the angle between $z$ directions of frame $i$ and $j$ which influences the overlap between two frames. The baseline length is defined as

$$t_{j,i} = \|\mathbf{t}_{w,i} - \mathbf{t}_{w,j}\|. \tag{5}$$

For each input image, two measurement frames in the former sequence are used to calculate the cost volume. The cost volume and the input image are then used by MVDepthNet to generate an inverse depth map. By inverting the values, we get full dense depth maps.

# 6. Implementation Details

## 6.1. MVDepthNet

The network is implemented on PyTorch [21] and trained from scratch using Adam [22] with $\beta_1 = 0.9$ and $\beta_2 = 0.999$. Although the MVDepthNet is designed to support multiple images, the network works by estimating inverse depth maps from reference images and constructed cost volumes thus do not know the number of measurement images. Exploiting this feature, we train the network using two-image pairs and apply it to two-view or multi-view depth estimation. All the images are normalized using the mean and variation of the dataset before any operation.

The loss is defined as the sum of the average L1 loss of estimated inverse depth maps at different resolutions compared with ground truth inverse depth maps:

$$L = \sum_{s=0}^{3} \frac{1}{n_s} \sum_{i} |\xi_{si} - \frac{1}{\hat{d}_{si}}|, \tag{6}$$

where $\xi_{si}$ is the estimated inverse depth at scale $s$ and $\hat{d}_{si}$ is the corresponding ground truth depth. $n_s$ is the number of pixels that have valid depth measurements from the dataset at scale $s$.

MVDepthNet is trained using $N_d = 64$ sampled depth value with a mini-batch size of $8$. The network is trained for $1140k$ iterations. For the first $300k$ iterations, the learning rate is set to $1e$-4 and divided by $2$ when the error plateaus in the remaining iterations. All the images for training and testing are scaled to $320 \times 256$ but without cropping.

## 6.2. Monocular Depth Estimation System

The trained network is applied to a real-time monocular depth estimation system that estimates depth maps of an image sequence. The threshold for measurement frames selection is a view angle change of $15$ degrees and baseline translation of $0.3$ meters.

# 7. Experiments

We first analyze the importance of the cost volume and the reference image in the ablation studies. The proposed geometric data augmentation is also studied, and we show that MVDepthNet can be easily extended to multiple images. MVDepthNet and the monocular depth estimation system are then compared with state-of-the-art open source methods. All the experiments are conducted on a workstation with a Nvidia TITAN-Xp, an i7-7700 CPU and 47 GB memory.

## 7.1. Datasets

Although the cost volume structure eases the network from searching corresponding pixels among images, a large dataset is needed to help the network learn the relationship between inverse depth maps and given information. Similar to DeMoN [7], we adopt multiple open source datasets. The training datasets include SUN3D [23], TUM RGB-D [24], MVS datasets (which contain images from [25], [26], [27], and [28]), SceneNN [29] and a synthetic dataset Scenes11 generalized in [7]. We select $436,928$ pairs for training and $13,394$ pairs for testing from the datasets. The selection strategy is similar to that of DeMoN [7] but stricter. To be specific, at least $70\%$ of the pixels in the reference image should be visible in the measurement image, and the average photo-consistency error between image pairs should be less than $80$. Except TUM RGB-D [24] and SceneNN [29], all training and test data are selected from the corresponding datasets used in DeMoN [7]. Six sequences (*freiburg1_desk*, *freiburg2_flowerbouquet*, *freiburg2_pioneer_slam2*, *freiburg2_xyz*, *freiburg3_cabinet*, and *freiburg3_nostructure_texture_far*) from the TUM RGB-D [24] are randomly selected to test our MVDepthNet-based monocular dense mapping system in Section 7.5, and others are used as the training data.

### 7.2. Error Metrics

Four measures are used: (1) L1-rel, (2) L1-inv, (3) sc-inv following DeMoN [7] and (4) correctly estimated depth percentage (C. P.) following CNN-SLAM [19]. To be specific,

$$\text{L1-rel} = \frac{1}{n}\sum_i \frac{|d_i - \hat{d}_i|}{\hat{d}_i}, \quad (7)$$

$$\text{L1-inv} = \frac{1}{n}\sum_i \left|\frac{1}{d_i} - \frac{1}{\hat{d}_i}\right|, \quad (8)$$

$$\text{sc-inv} = \sqrt{\frac{1}{n}\sum_i z_i^2 - \frac{1}{n^2}\left(\sum_i z_i\right)^2}, \quad (9)$$

where $d_i = 1/\xi_i$ is the estimated depth value, $\hat{d}_i$ is the corresponding ground truth value, and $z_i = \log d_i - \log \hat{d}_i$. C. P. represents the percentage of depth estimations in depth maps whose relative error is within $10\%$.

### 7.3. Ablation Studies

In the following sections, the importance of the cost volume and the reference image in MVDepthNet are studied. The benefit of geometric data augmentation and using more than two images are also shown in the ablation studies.

**The Effect of the Cost Volume.** Here, we analyze the performance of MVDepthNet given different numbers of sampled depths. Three networks with $N_d = 64, 32, 16$ sampled depth values are used (referred to as *MVDepthNet-64*, *MVDepthNet-32*, and *MVDepthNet-16* respectively). *MVDepthNet-0* stands for the approach where two input images are simply stacked as FlowNetSimple [13] and no cost volume is used. The number of convolutional layers in the encoder and decoder are adjusted so that networks have similar parameter numbers. These networks are trained using the same datasets and data augmentation for $400k$ steps.

| | Param. Num. | FPS | L1-rel | L1-inv | sc-inv | C. P. (%) |
|---|---|---|---|---|---|---|
| MVDepthNet-0 | 31.7M | 90.5 | 0.284 | 0.188 | 0.244 | 21.73 |
| MVDepthNet-16 | 31.8M | 69.4 | 0.171 | 0.101 | 0.197 | 42.16 |
| MVDepthNet-32 | 32.5M | 39.7 | 0.163 | 0.098 | 0.193 | 43.46 |
| MVDepthNet-64 | 33.9M | 25.0 | 0.155 | 0.097 | 0.190 | 43.95 |

Table 2. The performance using different sizes of cost volumes. Param. Num. is the number of parameters in the network. FPS measures the time including uploading data to the GPU and downloading results.

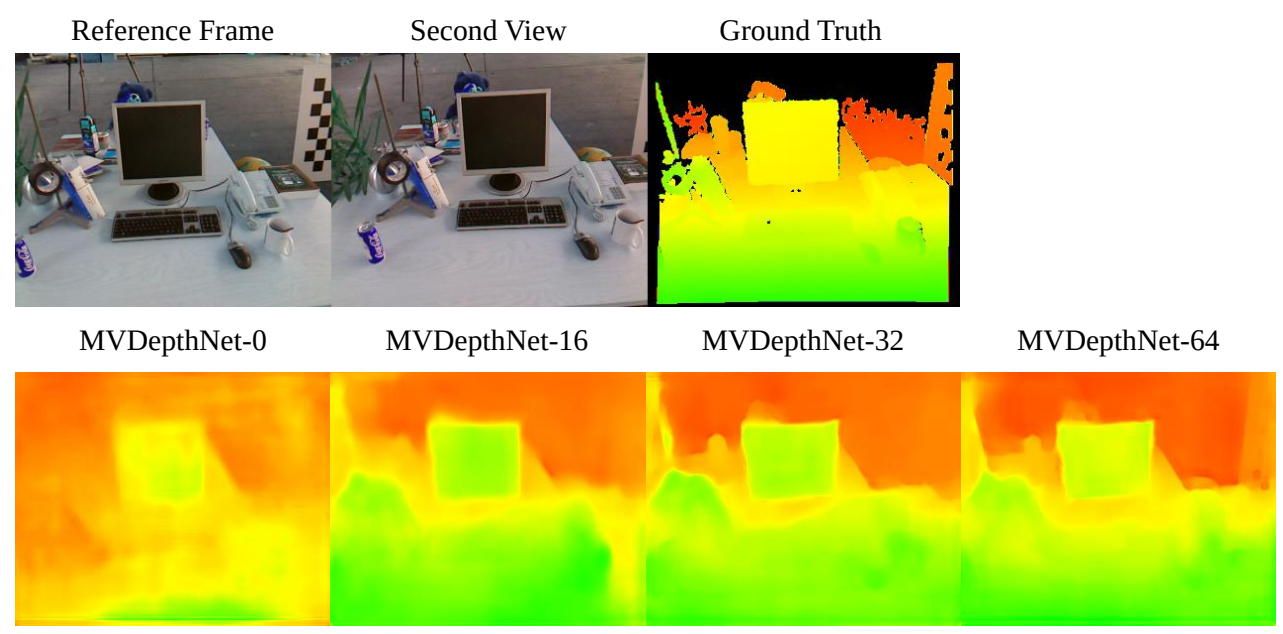

Figure 4. Qualitative results using different sizes of cost volumes. MVDepthNet-0 stands for simple stacking images and has difficulty estimating depth maps. Obviously, cost volumes effectively help the network to estimate depth maps. As more depth values are sampled, the network can estimate depth maps with more details and accuracy.

Table 2 and Figure 4 show the performance of MVDepthNet using different numbers of sampled depths. The constructed cost volume can effectively encode multiview information and help the network to estimate depth maps regardless of the camera motion and intrinsic parameters. Even a cost volume with $N_d = 16$ sampled depth values can help the network double the correctly estimated percentage compared with MVDepthNet-0 where two images are stacked as FlowNetSimple [13]. MVDepthNet-64 generates the most accurate depth estimation that L1-inv drops by half and C.P. doubles comparing with MVDepthNet-0. Although these networks have almost the same parameter numbers, FPS drops mainly due to the time to construct the cost volume.

**The Effect of the Reference Image.** *MVDepthNet-cost* is used to study the importance of the reference image in depth estimation. In MVDepthNet-cost, the reference image is set to zero and only the cost volumes are used in MVDepthNet. As shown in Table 3, MVDepthNet relies mostly on multiview observations and the performance only drops by $3\%$ to $5\%$ . Figure 5 shows the difference of one depth map generalized by MVDepthNet and MVDepthNet-cost. With the cost volume, the depth map is mostly accurate and works well in textureless regions (*e.g.*, the monitor). The reference image helps the network to smooth depth maps on surfaces and find object edges (*e.g.*, the bottom desk edge, and the plants).

| | L1-rel | L1-inv | sc-inv | C. P. (%) |
|---|---|---|---|---|
| MVDepthNet | 0.127 | 0.076 | 0.158 | 51.40 |
| MVDepthNet-cost | 0.133 | 0.079 | 0.164 | 50.13 |

Table 3. Performance of MVDepthNet using only cost volumes on all test images.

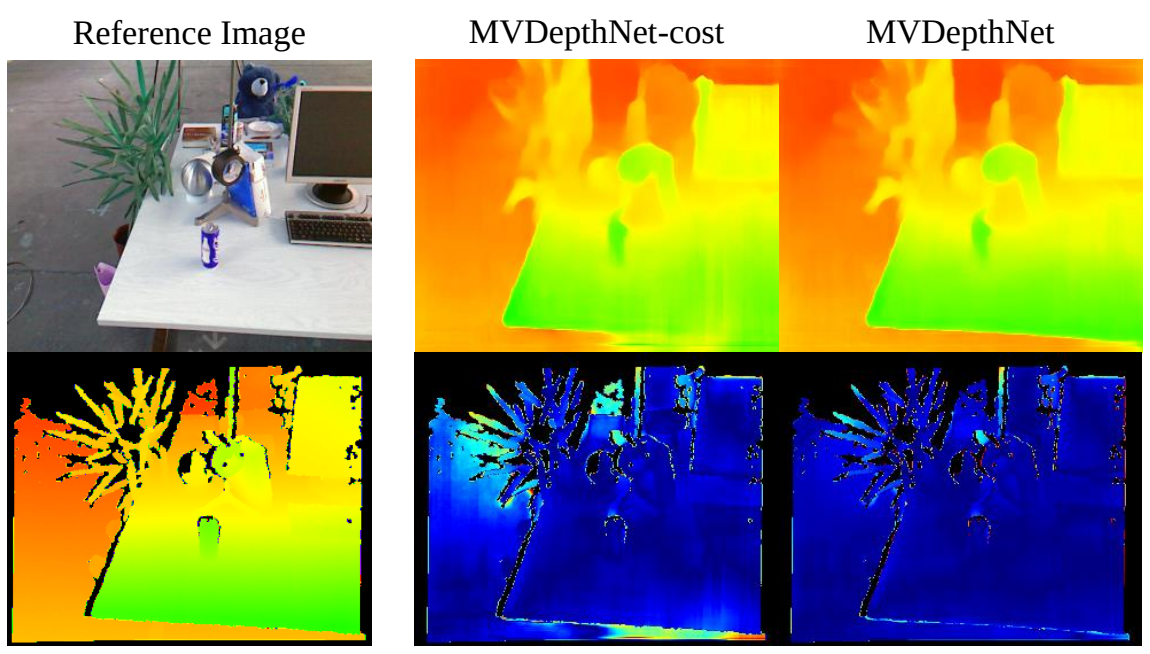

Figure 5. Depth maps when the reference image is not shown to the network. Left: the reference image and the corresponding ground truth depth. Right: the estimated depth maps and corresponding L1-inv error maps. Error maps are JET color-coded. MVDepthNet-cost can generate high quality depth estimations using only the cost volume. The reference image can help MVDepthNet smooth depth maps on surfaces and be accurate on edges.

**The Effect of Geometric Data Augmentation.** Geometric data augmentation is proposed to train the network with limited datasets while maintaining multiview geometry constraints. Here, we show that the augmentation can help the network estimate depth maps in better quality when images are taken in angles that are not covered by the training data, such as upside down. *MVDepthNet-NoGeo* is trained using the same settings as MVDepthNet-64 but without geometric data augmentation. The performance of these two networks on flipped test images is shown in Table 4 and Fig. 6. As demonstrated, geometric data augmentation helps the network to estimate scenes that are not covered in the training datasets, such as the ceiling.

| Flip Direction | Network | L1-rel | L1-inv | sc-inv | C. P.(%) |
|---|---|---|---|---|---|
| horizonal | MVDepthNet-64 | 0.153 | 0.095 | 0.188 | 44.65 |
| | MVDepthNet-NoGeo | 0.147 | 0.091 | 0.188 | 47.58 |
| vertical | MVDepthNet-64 | 0.152 | 0.094 | 0.188 | 44.78 |
| | MVDepthNet-NoGeo | 0.205 | 0.139 | 0.274 | 37.11 |

Table 4. Comparison of networks trained using geometric data augmentation. With geometric data augmentation, the network performs consistently well when images are vertical flipped.

**Using Multiple Images** One of the benefits of the proposed cost volume is that MVDepthNet can process more than two images within one inference. These images are taken from different viewpoints and can minimize the depth ambiguity of textureless regions. Using Equation 2, multiple images are encoded into the cost volume and used as the input of MVDepthNet. Figure 7 shows an example where MVDepthNet takes up to 6 images (including the reference image) as the input. As more images are encoded into the cost volume, MVDepthNet performs better. The depth estimation of both textureless regions (*e.g.*, table) and distant areas benefits from multiview observations.

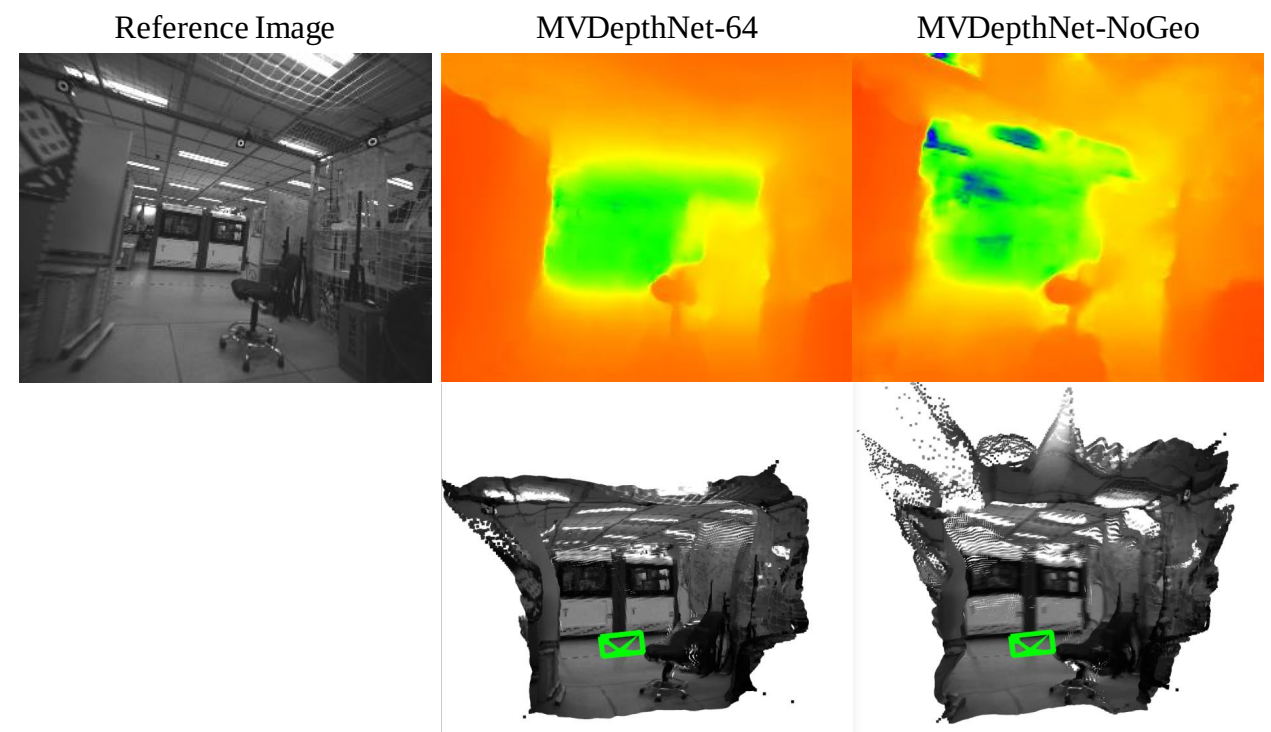

Figure 6. Performance comparison when geometric data augmentation is not used. The estimated depth maps are shown in the first row, and the corresponding projected point clouds are shown in the second row. Camera poses are marked in green. MVDepthNet-NoGeo cannot handle the ceiling and distant objects well.

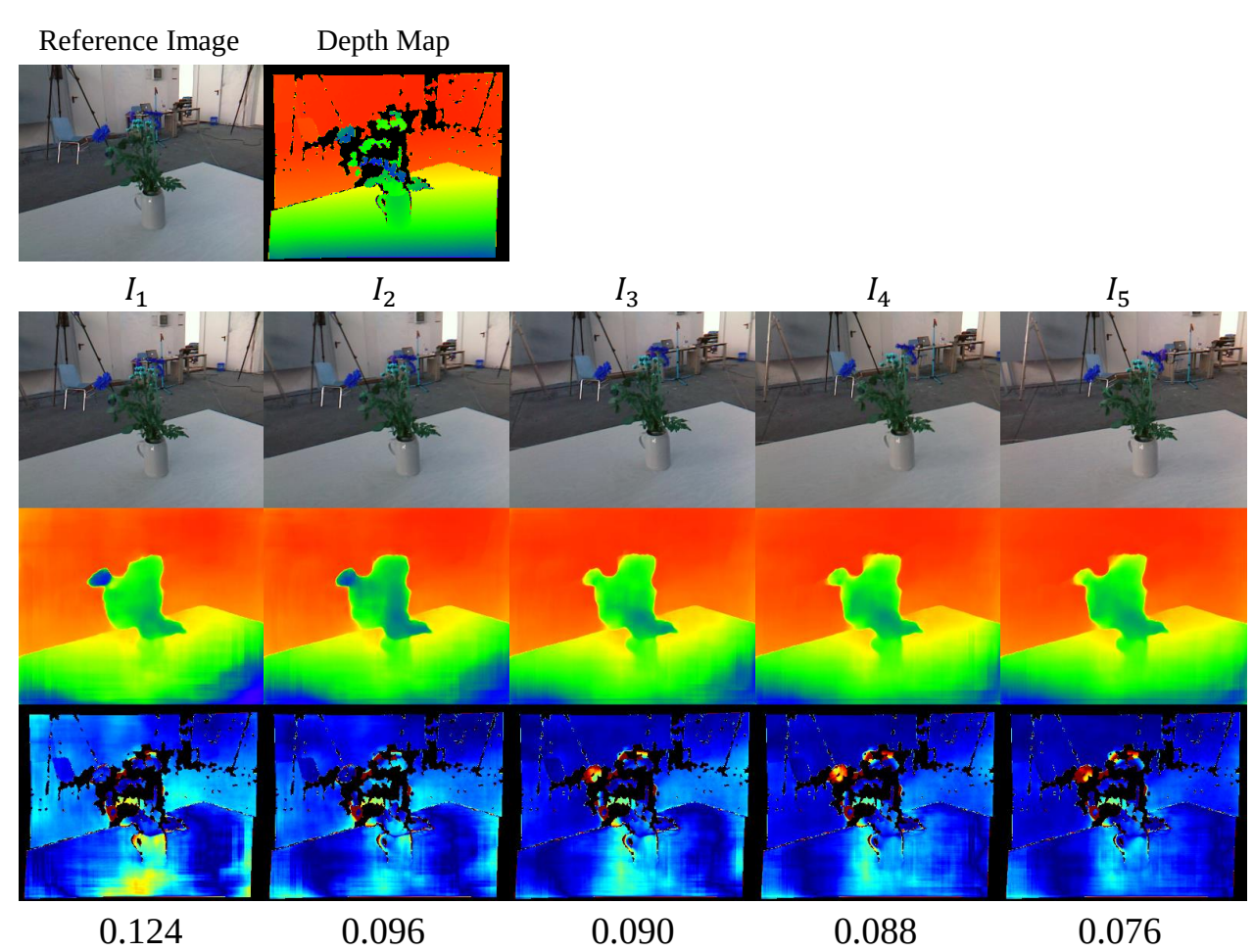

Figure 7. MVDepthNet can take more than two images as input. The first row is the reference image and the corresponding depth map. The second-row images are measurement images. Depth maps are estimated using all the former measurement images. For example, the third depth map is estimated using the reference image and three measurement images ($I_1$, $I_2$ and $I_3$). L1-inv error maps and L1-inv errors are shown in the fourth row and bottom. The more multiview images that are taken, the more accurate the depth maps estimated by MVDepthNet are.

### 7.4. Comparison Using Image Pairs

MVDepthNet is compared with state-of-the-art methods DeMoN [7], FCRN-DepthPrediction [9] (referred to as *FCRN*) and DeepMVS [8].

DeMoN [7] takes two images as the input and generates motion and depth estimations. To test DeMoN [7], images are cropped to match required intrinsic parameters and scaled to $256 \times 192$ according to its default settings. Since the depth estimation is based on unit translation, we scale the output using ground truth camera translation as Huang *et al*. [8].

FCRN [9] is a state-of-the-art depth prediction method and is the core part of CNN-SLAM [19]. We use the model trained on the NYU Depth v2 dataset [30]. Following CNN-SLAM [19], to improve its results, we further adjust the predicted depth maps using the ratio between the current camera focal length and the focal length in its training datasets.

During the evaluation of DeepMVS [8], images are scaled to $320 \times 256$ to speed up the processing time. DeepMVS [8] takes 3.03 seconds to estimate one depth map and MVDepthNet is over 70 times faster processing the same resolution images. The estimated depth maps are refined using Dense-CRF [17] before evaluation.

| Dataset | Method | L1-rel | L1-inv | sc-inv | C.P. (%) |
|---|---|---|---|---|---|
| SUN3D | MVDepthNet | **0.139** | **0.066** | **0.154** | **45.11** |
| | DeMoN | 0.211 | 0.115 | 0.171 | 27.67 |
| | DeepMVS | 0.362 | 0.113 | 0.301 | 33.22 |
| | FCRN | 0.219 | 0.116 | 0.195 | 38.82 |
| TUM RGB-D | MVDepthNet | **0.130** | **0.069** | **0.180** | **48.11** |
| | DeMoN | - | - | - | - |
| | DeepMVS | - | - | - | - |
| | FCRN | 0.314 | 0.211 | 0.307 | 21.99 |
| MVS | MVDepthNet | 0.167 | 0.086 | 0.238 | 56.25 |
| | DeMoN | 0.329 | 0.115 | 0.272 | 33.17 |
| | DeepMVS | **0.129** | **0.060** | **0.228** | **61.24** |
| | FCRN | 0.423 | 0.215 | 0.298 | 18.04 |
| SceneNN | MVDepthNet | **0.123** | **0.082** | **0.138** | **53.55** |
| | DeMoN | 0.319 | 0.195 | 0.183 | 25.20 |
| | DeepMVS | 0.199 | 0.117 | 0.207 | 43.17 |
| | FCRN | 0.397 | 0.220 | 0.206 | 20.92 |
| Scenes11 | MVDepthNet | **0.120** | 0.035 | 0.305 | **75.84** |
| | DeMoN | 0.243 | **0.029** | **0.251** | 38.93 |
| | DeepMVS | 0.276 | 0.039 | 0.453 | 55.70 |
| | FCRN | 0.769 | 0.492 | 0.616 | 1.47 |

Table 5. Comparison with state-of-the-art methods. MVDepthNet performs consistently well in all datasets and outperforms other methods in most of the datasets and error metrics. DeMoN [7] and DeepMVS [8] are not tested on TUM RGB-D [24] since some images are used in their training datasets.

Different from DeMoN[7] which only evaluates pixels that are visible in both images, we evaluate all pixels that have corresponding ground truth values because all methods produce fully dense depth maps. Table 5 shows that MVDepthNet outperforms other methods in most of the datasets and error metrics. Please see the supplementary material for detailed qualitative comparisons. We notice that MVDepthNet performs the best in Scenes11 in terms of L1-rel, L1-inv and C.P. On the contrary, FCRN [9] performs the worst on this dataset. This can be explained by the fact that MVDepthNet relies mostly on multiview observation of a pixel, thus is insensitive to scene contents. However, FCRN [9] predicts depth maps based on learned priors, so that cannot work well in unseen scenes. MVDepthNet outperforms DeMoN [7] in most of the cases because input camera poses limit the corresponding pixels searching space from a 2D image to a 1D epipolar line. Given the relative camera poses and constructed cost volume, MVDepthNet can search and triangulate pixels efficiently in high precision. DeepMVS [8] performs the best in MVS dataset due to the high-quality images and camera poses estimation. However, DeepMVS [8] only estimates discrete depth labels and cannot handle Scenes11 [7] dataset because of many occlusions and textureless objects.

### 7.5. Comparison Using Image Sequences

The monocular depth estimation system is compared with two state-of-the-art mapping systems–REMODE [15] and VI-MEAN [31]. REMODE [15] updates the depth of each pixel in an outlier-robust way and smooths depth maps using a total variation optimization. VI-MEAN [31] solves depth maps by a semiglobal optimization.

Six sequences from the TUM RGB-D dataset [24] and four sequences from the ICL-NUIM dataset [32] are used to evaluate the estimated depth maps. The results are shown in Table 6. The detailed performance on each sequence and qualitative comparisons can be found in the supplementary material. Our method performs consistently well and is the best in all real-world experiments. However, VI-MEAN [31] estimates depth maps most accurately in synthetic datasets. This shows that perfect camera poses, and light conditions are important for traditional methods, and our learning-based method is more suitable for real-world data where noise exists in images and poses.

| Dataset | Method | L1-rel | L1-inv | C.P. (%) | Density (%) |
|---|---|---|---|---|---|
| TUM RGB-D | OURS | **0.122** | **0.069** | **50.81** | **100.00** |
| | VI-MEAN | 0.892 | 0.230 | 18.53 | 72.21 |
| | REMODE | 0.916 | 0.283 | 8.67 | 34.09 |
| ICL-NUIM | OURS | 0.144 | 0.072 | 57.68 | **100.00** |
| | VI-MEAN | **0.092** | **0.036** | **75.70** | 95.69 |
| | REMODE | 0.227 | 0.227 | 42.23 | 60.72 |

Table 6. Evaluation of monocular depth estimation methods. Our method performs consistently well on two datasets and outperforms the others on the real-world dataset by a wide margin.

## 8. Conclusion

In this paper, we propose MVDepthNet, a network that generates depth maps using multiview images and corresponding camera poses from a monocular camera. The cost volume is used to encode multiview observations so that MVDepthNet can find pixel correspondence and triangulate depth values efficiently. Thanks to the cost volume layer, our method achieves the flexibility that can be applied to more than two images and cameras with intrinsic parameters that are different from that of the training datasets. Geometric data augmentation is proposed to train the network using limited datasets. A monocular dense mapping system is also developed using MVDepthNet that continuously estimates depth maps from image-pose sequences. The performance of MVDepthNet and the monocular mapping system is demonstrated using a variety of datasets and compared with other state-of-the-art methods.

# MVDepthNet: Real-time Multiview Depth Estimation Neural Network – Supplementary Material –

## 1. Cost Volume Computation

Although the matching cost in the cost volume is defined per-pixel, the cost volume can be calculated for all pixels at a certain distance by perspective warping the reference image. Given warp matrix $\mathbf{P}$, the pixel with coordinate $(u, v)$ is transformed into a new image at $(u', v')$:

$$\begin{bmatrix} u' \\ v' \\ 1 \end{bmatrix} = \lambda(\mathbf{P} \begin{bmatrix} u \\ v \\ 1 \end{bmatrix}) \tag{1}$$

where $\lambda(\cdot)$ is the normalize function that $\lambda([x, y, z]^T) = [x/z, y/z, 1]^T$. Given the camera intrinsic matrix $\mathbf{K}$, relative camera pose $\mathbf{T}_{m,r} = \mathbf{T}_{w,m}^{-1}\mathbf{T}_{w,r}$ and sampled depth $d$, the pixel in the reference image $\mathbf{u}_r$ can be projected into the measurement frame as $\mathbf{u}_m$:

$$\mathbf{u}_m = \lambda(d\mathbf{K}\mathbf{R}_{m,r}\mathbf{K}^{-1}\mathbf{u}_r + \mathbf{K}\mathbf{t}_{m,r}), \tag{2}$$

where $\mathbf{R}_{m,r}$ and $\mathbf{t}_{m,r}$ is the rotation matrix and the translation vector in $\mathbf{T}_{m,r}$. Since $\mathbf{u}_r = [u, v, 1]^T$, Equation 2 can be written as:

$$\mathbf{u}_m = \lambda((d\mathbf{K}\mathbf{R}_{m,r}\mathbf{K}^{-1} + \begin{bmatrix} 0 & 0 & \mathbf{K}\mathbf{t}_{m,r}(0) \\ 0 & 0 & \mathbf{K}\mathbf{t}_{m,r}(1) \\ 0 & 0 & \mathbf{K}\mathbf{t}_{m,r}(2) \end{bmatrix})\mathbf{u}_r), \tag{3}$$

and we have

$$\mathbf{P} = d\mathbf{K}\mathbf{R}_{m,r}\mathbf{K}^{-1} + \begin{bmatrix} 0 & 0 & \mathbf{K}\mathbf{t}_{m,r}(0) \\ 0 & 0 & \mathbf{K}\mathbf{t}_{m,r}(1) \\ 0 & 0 & \mathbf{K}\mathbf{t}_{m,r}(2) \end{bmatrix}. \tag{4}$$

The cost for all pixels at the sampled depth $d$ comparing with $I_m$ can be calculated by perspective warping the reference image $I_r$ into $I'_r$ using $\mathbf{P}$ and computing the absolute difference between $I'_r$ and $I_m$.

## 2. Datasets

A large number of image pairs are used to train the network. Table 1 shows the details of the datasets used for training and testing. Most of the real datasets, including SUN3D [1], TUM RGB-D [2], and SceneNN [3], are captured by RGB-D cameras so that the range of ground truth depth is limited. Although Scenes11 [4] provides perfect poses and depth maps with much larger ranges, it is not photorealistic. The depth range is cropped to $0.5$ m to $50.0$ m to meet the coverage of sampled depth values.

## 3. Experiments

We show additional evaluation details and qualitative comparisons of MVDepthNet and other state-of-the-art methods.

### 3.1. Model Saliency

In this section, we show that MVDepthNet estimates depth maps using multiview observations of large regions and focuses on related objects. Saliency maps [5] are shown in Figure 1 that pixels are colored by the influence to the depth estimation of the selected pixels. The influence value is measured by the gradient of the input images with respect to the output depth estimation of the pixels. As shown in the figure, MVDepthNet utilizes information of large areas (*e.g.*, in 'ground' and 'fine structure') and focuses on related objects (*e.g.*, in 'desk' and 'chair') to estimate the depth of a pixel.

### 3.2. Comparison Using Image Pairs

MVDepthNet is compared with DeMoN [4][1], FCRN-DepthPrediction [6][2](referred to as FCRN) and DeepMVS [7][3] using two-image pairs from a variety of datasets.

The qualitative results including the estimated inverse depth maps and L1-inv error maps are shown from Figure 2 to Figure 6. As shown in the figures, FCRN [6] predicts indoor scenes well but cannot generalize to outdoor environments. Both MVDepthNet and DeMoN [4] estimates smooth and accurate depth maps due to the usage of multiview observations. MVDepthNet takes camera poses and thus can search and triangulate the depth values more precisely. On the contrary, DeMoN [4] estimates depth maps together with ego-motion. Without the information of camera poses, DeMoN [4] has to search pixel correspondences in 2D images thus needs more time. Since the relative pose is not perfectly estimated, the depth values cannot be triangulated as accurately as MVDepthNet which encodes the information in the cost volume. DeepMVS [7] works well in the MVS dataset which contains high-quality images and camera poses. However, since the training loss is defined on pixel classification, DeepMVS [7] cannot handle occlusion well such as the chair sequence in Figure 4.

[1] https://github.com/lmb-freiburg/demon
[2] https://github.com/iro-cp/FCRN-DepthPrediction
[3] https://github.com/phuang17/DeepMVS

| Datasets | Image Num. (train/test) | Perfect GT | Photorealistic | Mean Trans. | Mean Min Depth | Mean Max Depth |
|---|---|---|---|---|---|---|
| SUN3D | 91834/57 | No | Yes | 0.240 | 0.5 | 3.297 |
| TUM RGB-D | 23163/5750 | No | Yes | 0.191 | 0.5 | 6.201 |
| MVS | 12147/58 | No | Yes | 4.918 | 0.5 | 24.806 |
| SceneNN | 89465/7403 | No | Yes | 0.104 | 0.5 | 2.995 |
| Scenes11 | 220319/126 | Yes | No | 0.744 | 2.514 | 50.0 |

Table 1. Details of the training and testing datasets.

### 3.3. Comparison Using Image Sequences

| Sequence | Method | L1-rel | L1-inv | C.P. (%) | Density (%) |
|---|---|---|---|---|---|
| desk | OURS | **0.109** | **0.083** | **55.42** | **100.00** |
| | VI-MEAN | 0.623 | 0.197 | 28.35 | 78.29 |
| | REMODE | 1.314 | 0.319 | 7.53 | 30.78 |
| flowerbouquet | OURS | **0.160** | **0.073** | **43.18** | **100.00** |
| | VI_MEAN | 0.770 | 0.194 | 20.21 | 77.99 |
| | REMODE | 0.266 | 0.190 | 8.25 | 25.47 |
| pioneer slam2 | OURS | **0.161** | **0.070** | **36.46** | **100.00** |
| | VI-MEAN | 1.019 | 0.336 | 5.70 | 53.12 |
| | REMODE | 2.161 | 0.286 | 3.26 | 41.39 |
| xyz | OURS | **0.076** | **0.052** | **55.53** | **100.00** |
| | VI-MEAN | 0.564 | 0.181 | 23.27 | 79.94 |
| | REMODE | 0.801 | 0.633 | 10.30 | 48.28 |
| cabinet | OURS | **0.146** | **0.101** | **49.22** | **100.00** |
| | VI-MEAN | 1.526 | 0.299 | 13.62 | 69.70 |
| | REMODE | 0.421 | 0.176 | 6.03 | 22.86 |
| nostructure texture far | OURS | **0.079** | **0.036** | **65.06** | **100.00** |
| | VI-MEAN | 0.848 | 0.173 | 20.03 | 74.20 |
| | REMODE | 0.535 | 0.096 | 16.62 | 35.76 |
| lr kt0 | OURS | 0.173 | 0.080 | 61.34 | **100.00** |
| | VI-MEAN | **0.093** | 0.033 | **84.53** | 97.14 |
| | REMODE | 0.141 | **0.027** | 64.18 | 69.33 |
| lr kt1 | OURS | 0.129 | 0.069 | 54.58 | **100.00** |
| | VI-MEAN | **0.072** | **0.027** | **79.94** | 97.25 |
| | REMODE | 0.153 | 0.058 | 55.28 | 67.00 |
| of kt0 | OURS | 0.165 | 0.083 | 52.20 | **100.00** |
| | VI-MEAN | **0.121** | **0.048** | **64.34** | 91.80 |
| | REMODE | 0.122 | 0.432 | 28.85 | 53.65 |
| of kt1 | OURS | 0.108 | 0.054 | 62.60 | **100.00** |
| | VI-MEAN | **0.080** | **0.035** | **73.98** | 96.56 |
| | REMODE | 0.513 | 0.390 | 20.60 | 52.90 |

Table 2. Evaluation of monocular depth estimation methods. The first six sequences are from the TUM RGB-D dataset [2] and others are from the ICL-NUIM dataset [8]. Our method which uses MVDepthNet as the depth estimator outperforms the others in all TUM RGB-D sequences. On the contrary, VI-MEAN [9] performs the best in the ICL-NUIM dataset [8] in terms of accuracy because of the perfect camera poses and light conditions.

A monocular dense mapping system is built using MVDepthNet as the core solver. The system is compared with REMODE [10] and VI-MEAN [9] using sequences from TUM RGB-D dataset [2] and ICL-NIUM dataset [8]. we modified VI-MEAN [9] according to their successor work [11] that decouples visual odometry and mapping so that VI-MEAN [9] can use the ground truth camera poses. During the experiments, only valid depth estimations are evaluated. To be specific, masked pixels of VI-MEAN [9] and diverged pixels of REMODE [10] are not evaluated. All pixels in the depth maps of our method are evaluated. The performance of each method on each sequence can be found in Table 2. The snapshots of the estimated depth maps is shown in Figure 7 and Figure 8. Since REMODE [10] and VI-MEAN [9] estimate depth maps for selected keyframes, the reference images are not the same in the figures. As shown in the figures, MVDepthNet based mapping system works consistently on all datasets, but REMODE [10] and VI-MEAN [9] cannot estimate well on TUM RGB-D dataset in which images have noises from rolling shutter, exposure time changes and motion blur exists. On the contrary, REMODE [10] and VI-MEAN [9] estimate better depth maps on the ICL-NIUM dataset which contains perfect synthetic images and poses. Due to the 1D regularization, the depth maps from VI-MEAN [9] contain 'streaking' artifacts. And REMODE [10] does not deal with textureless regions well. Please find more comparison in the supplementary video.

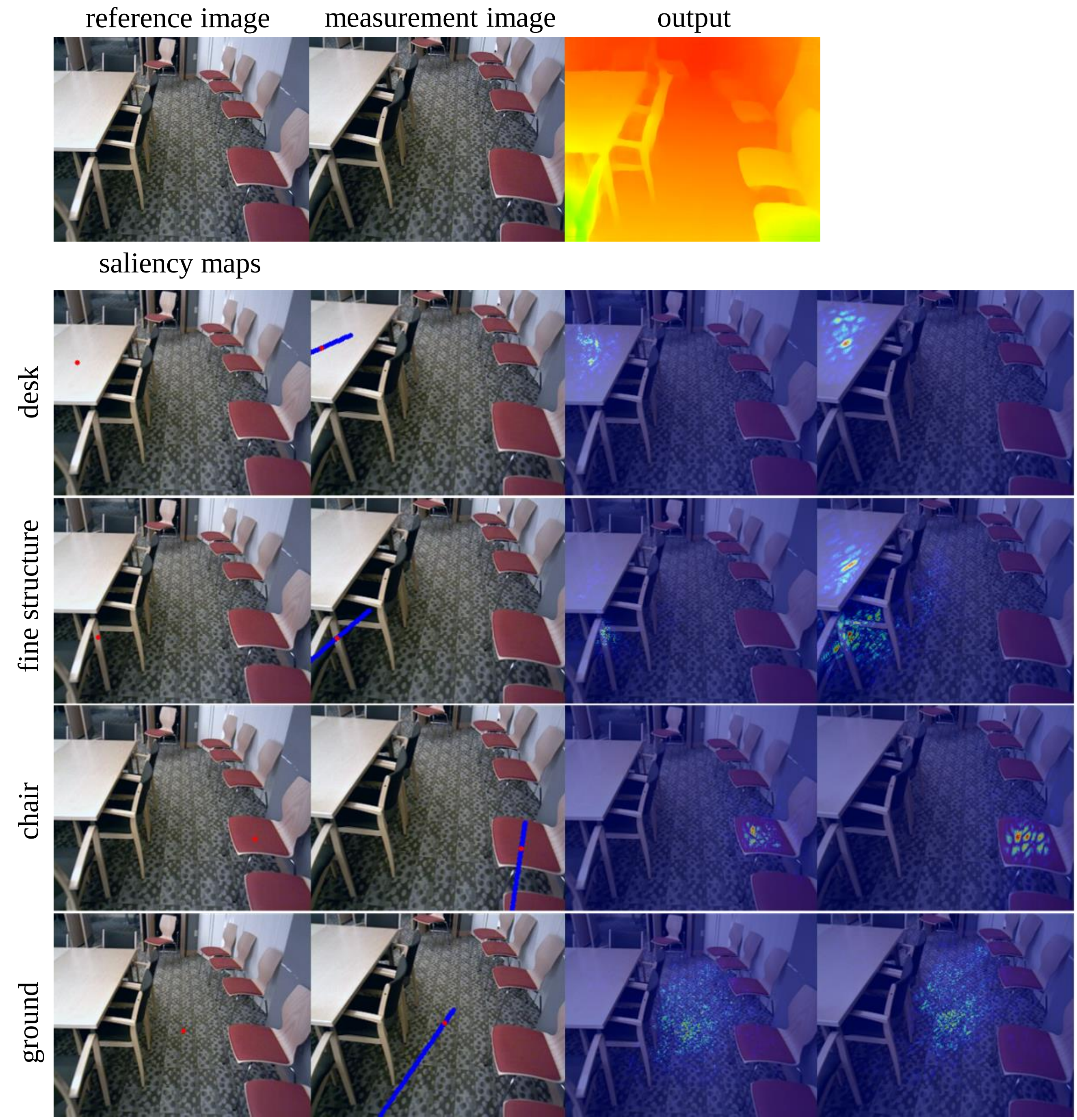

Figure 1. A visualization of saliency map. Top: the reference image, the measurement image and the depth estimation of MVDepthNet. Bottom: Selected points are marked in red in the reference images. Searched pixels are marked in blue in the measurement frames and the true correspondences are colored in red. Saliency maps are JET color-coded and overlap with images for visualization. Hot means high influence on the estimation of the selected pixels. As shown, MVDepthNet can use the information of large regions and focus on related objects.

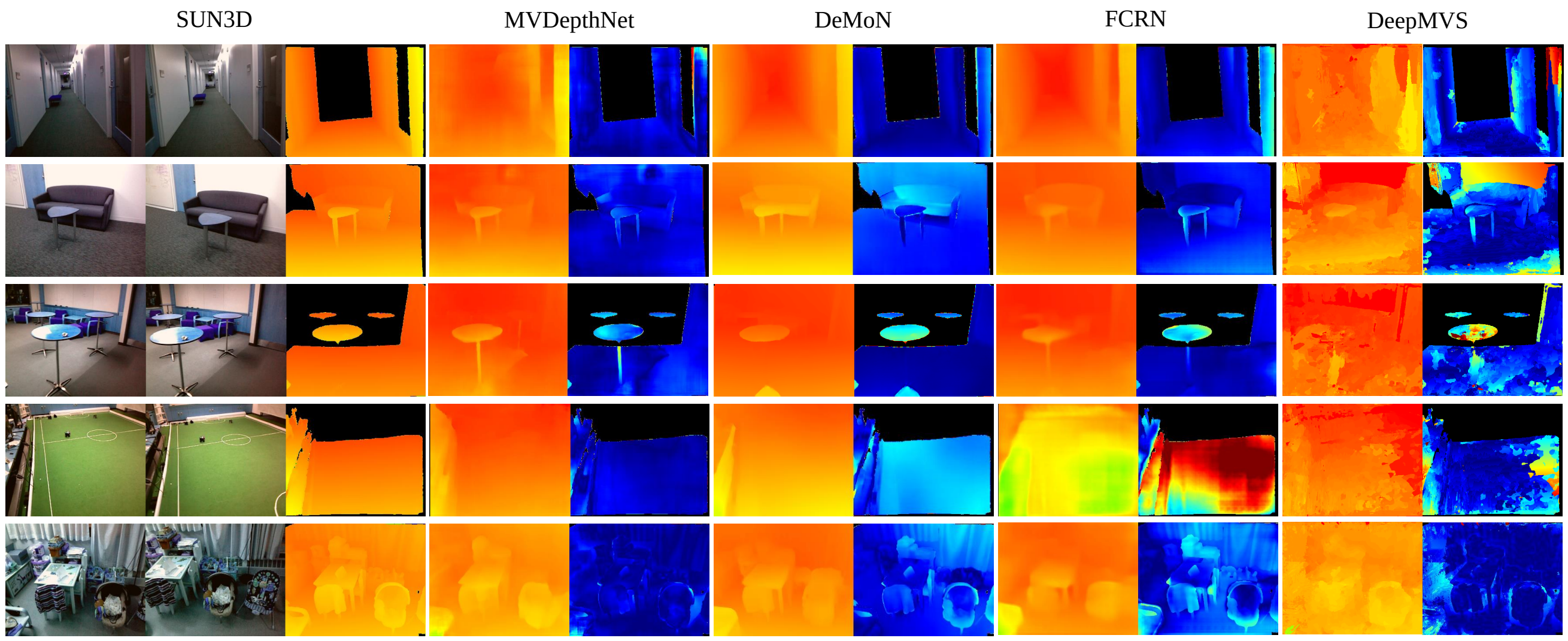

Figure 2. Qualitative results on SUN3D dataset. Left is the reference image, second view image and the 'ground truth' provided in the dataset. For every method, the estimated inverse depth maps and error maps are shown respectively.

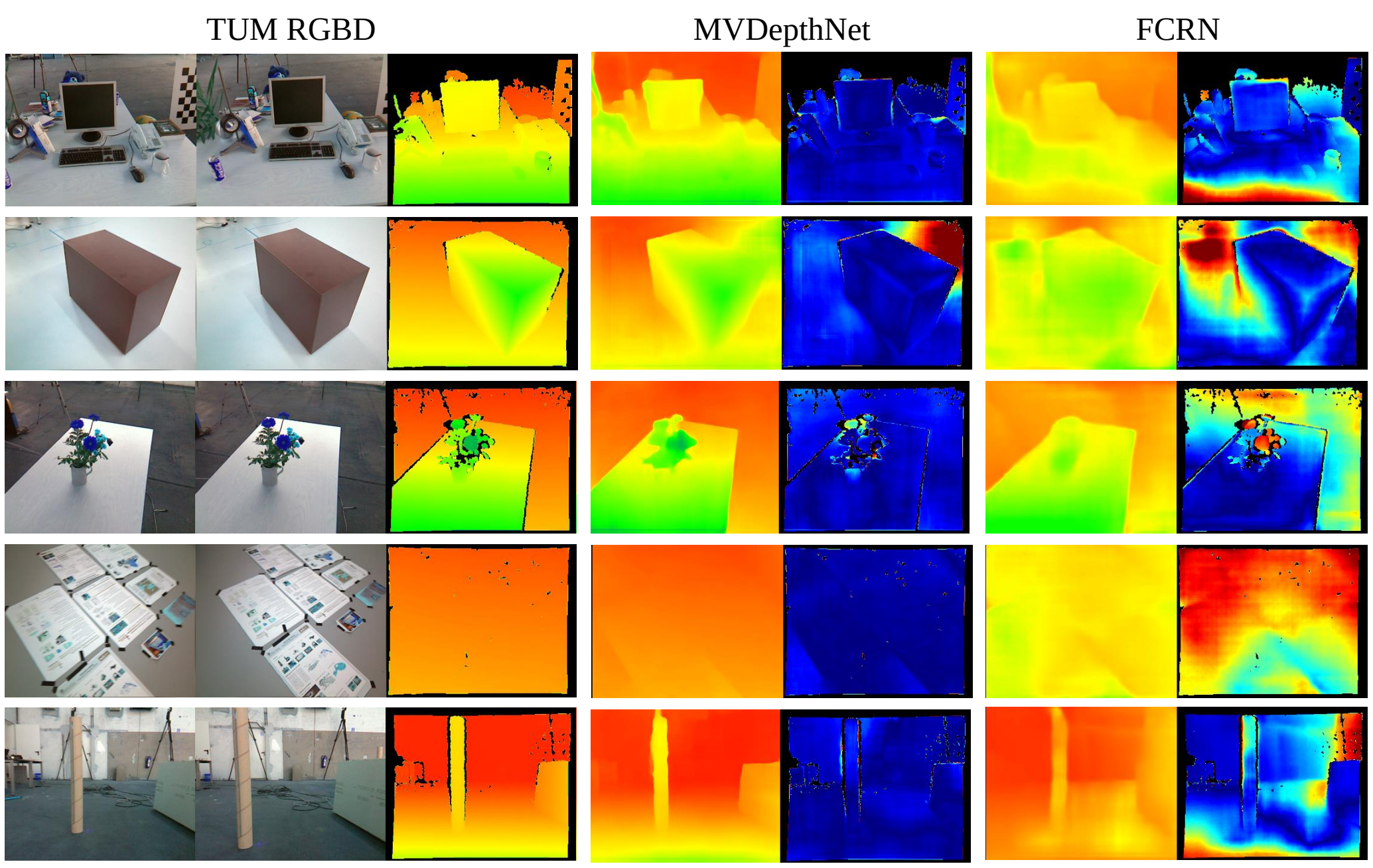

Figure 3. Qualitative results on TUM RGB-D dataset. DeMoN and DeepMVS are not compared using TUM RGB-D dataset since some of these images are used to train the networks.

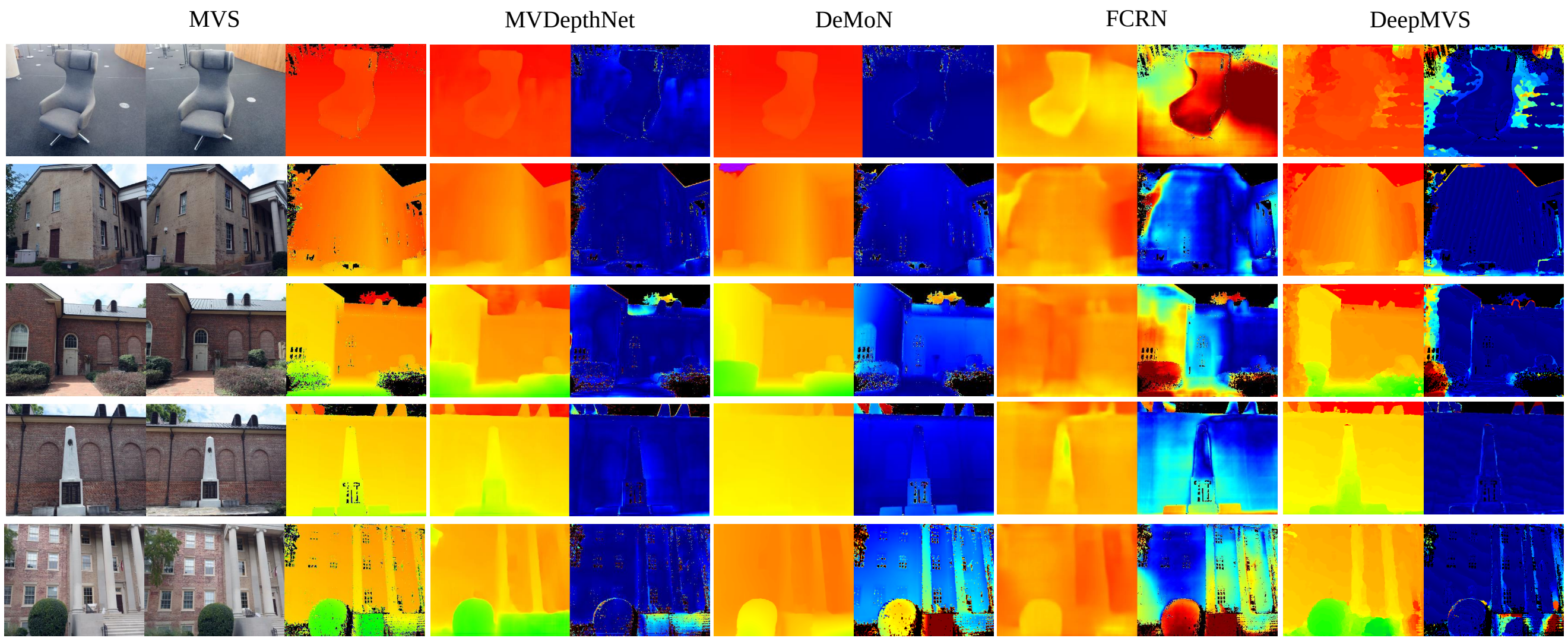

Figure 4. Qualitative results on MVS dataset.

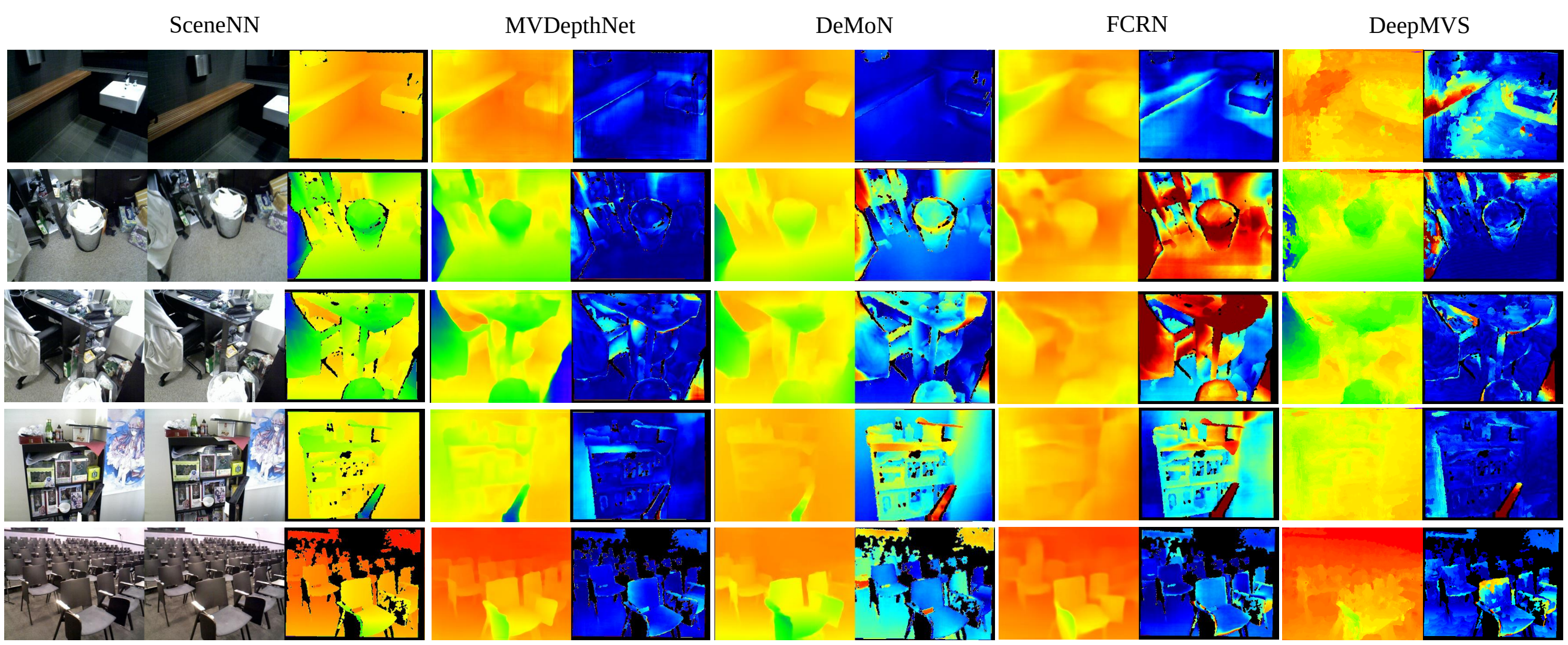

Figure 5. Qualitative results on SceneNN dataset.

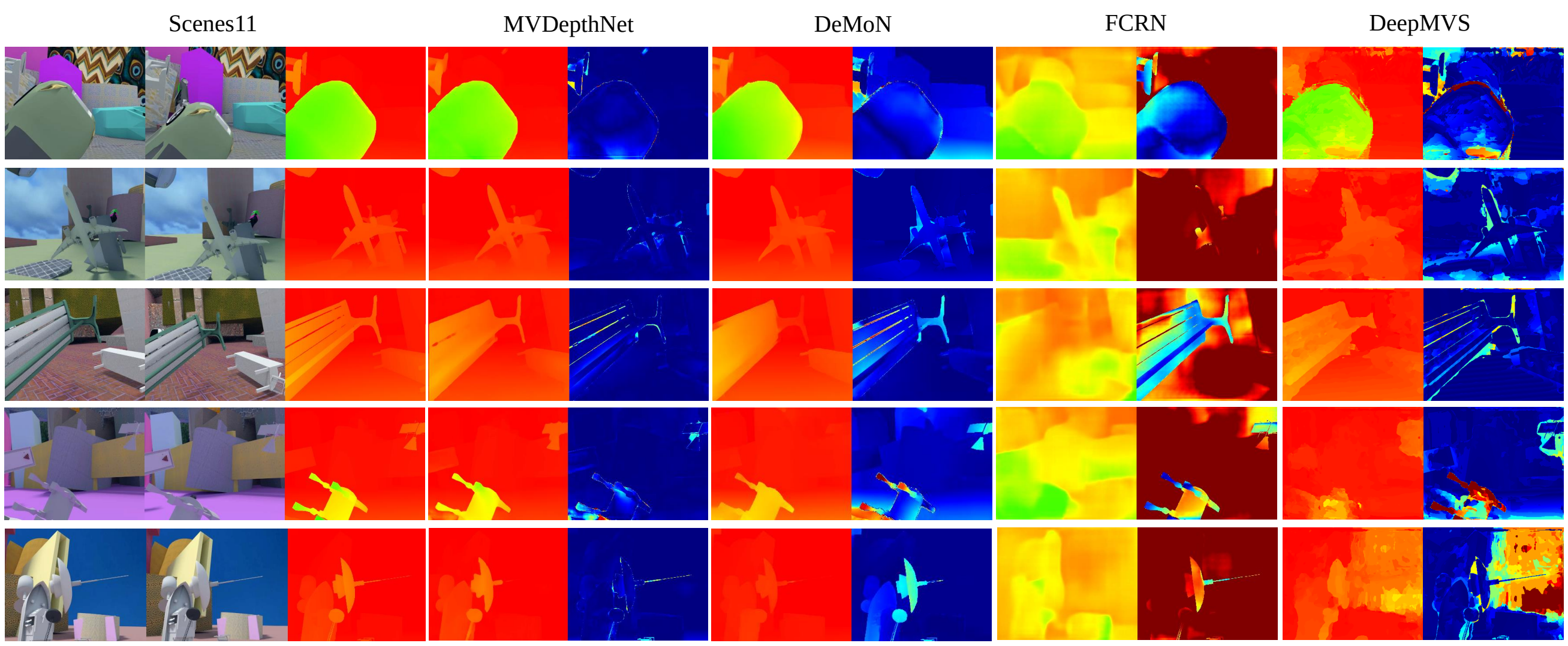

Figure 6. Qualitative results on Scenes11 dataset.

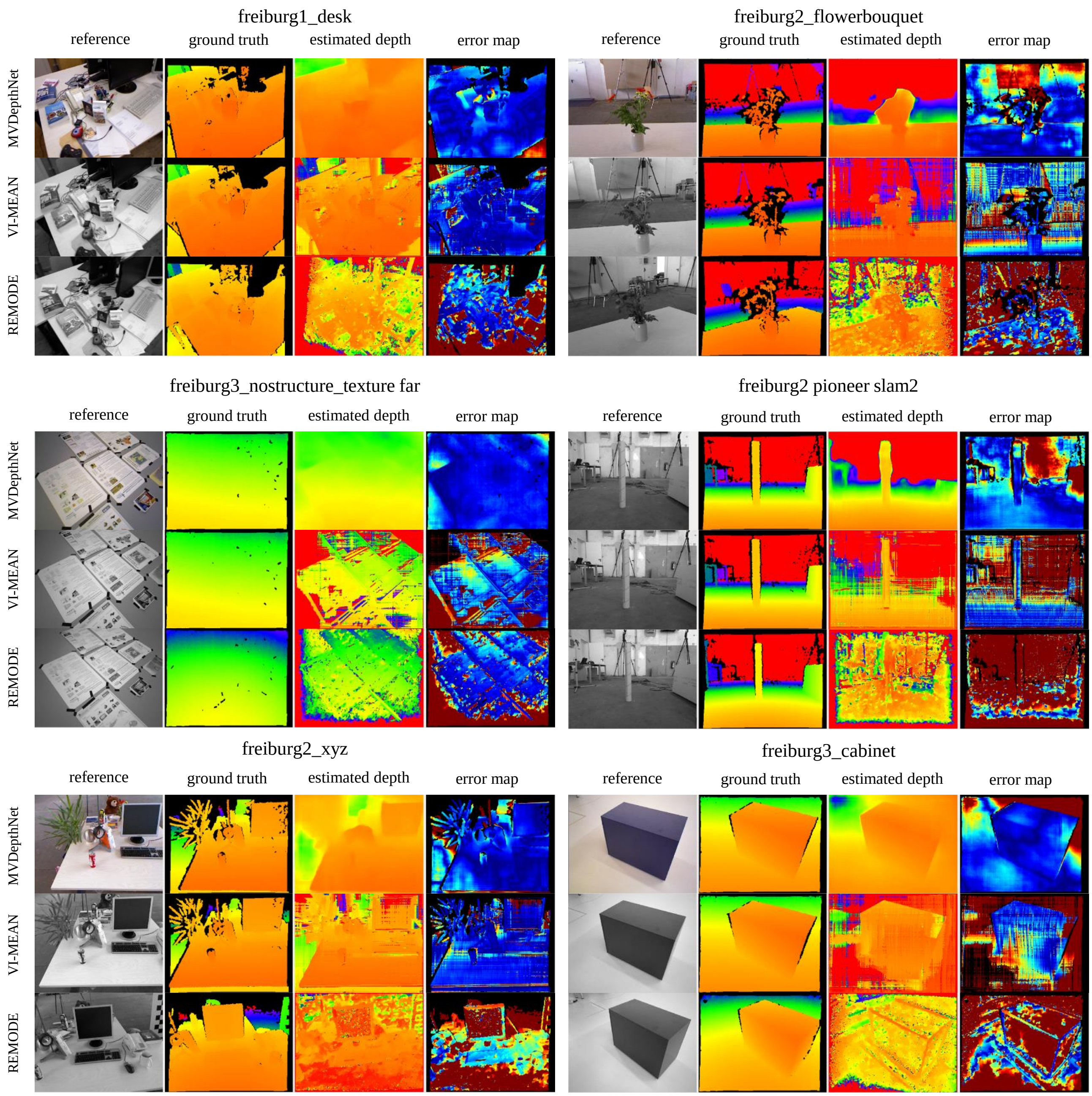

Figure 7. Qualitative results using image sequences from TUM RGB-D dataset.

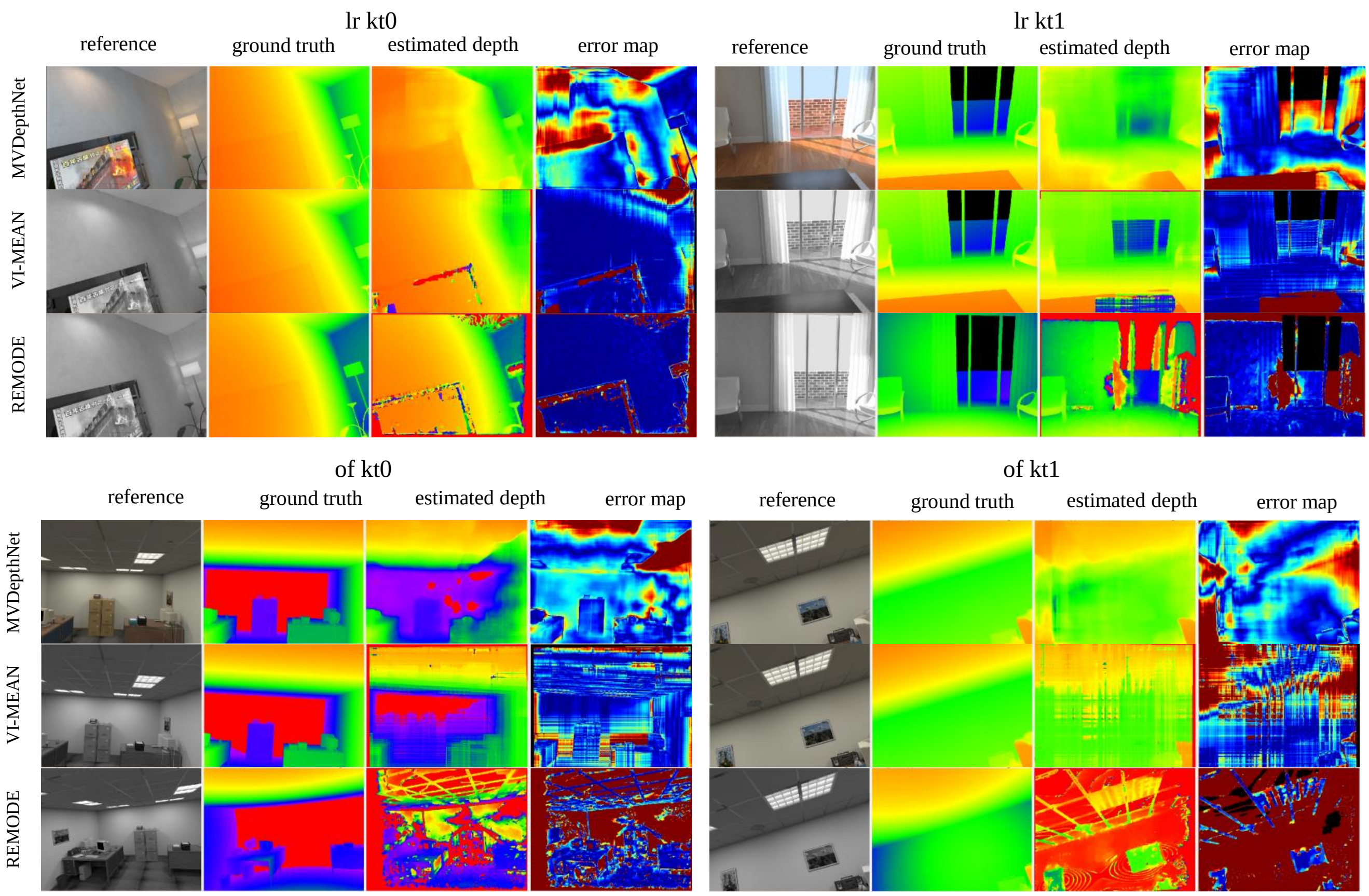

Figure 8. Qualitative results using image sequences from the ICL-NIUM dataset.